\documentclass[english]{article}
\usepackage[T1]{fontenc}
\usepackage[latin9]{inputenc}
\usepackage{float}
\usepackage{hyperref}
\usepackage{amsmath}
\usepackage{graphicx}
\usepackage{babel}
\begin{document}

\title{Fractional Max-Pooling}

\author{
Benjamin Graham\\
{\small Dept of Statistics, University of Warwick, CV4 7AL, UK}\\
{\small \tt b.graham@warwick.ac.uk}\\
}
\maketitle
\begin{abstract}
Convolutional networks almost always incorporate some form of spatial
pooling, and very often it is $\alpha\times\alpha$ max-pooling with
$\alpha=2$. Max-pooling act on the hidden layers of the network,
reducing their size by an integer multiplicative factor $\alpha$.
The amazing by-product of discarding 75\% of your data is that you
build into the network a degree of invariance with respect to translations
and elastic distortions. However, if you simply alternate convolutional
layers with max-pooling layers, performance is limited due to the
rapid reduction in spatial size, and the disjoint nature of the pooling
regions. We have formulated a \emph{fractional} version of max-pooling
where $\alpha$ is allowed to take non-integer values. Our version
of max-pooling is \emph{stochastic} as there are lots of different
ways of constructing suitable pooling regions. We find that our form
of fractional max-pooling reduces overfitting on a variety of datasets:
for instance, we improve on the state of the art for CIFAR-100 without
even using dropout.
\end{abstract}

\section{Convolutional neural networks}

Convolutional networks are used to solve image recognition problems.
They can be built by combining two types of layers:
\begin{itemize}
\item Layers of convolutional filters.
\item Some form of spatial pooling, such as max-pooling.
\end{itemize}
Research focused on improving the convolutional layers has lead to
a wealth of techniques such as dropout \cite{dropout}, DropConnect
\cite{DropConnect}, deep networks with many small filters\cite{multicolumndeep},
large input layer filters for detecting texture \cite{conf/nips/KrizhevskySH12},
and deeply supervised networks \cite{DeeplySupervisedNets}.

By comparison, the humble pooling operation has been slightly neglected.
For a long time $2\times2$ max-pooling (MP2 has been the default
choice for building convolutional networks. There are many reasons
for the popularity of MP2-pooling: it is fast, it quickly reduces
the size of the hidden layers, and it encodes a degree of invariance
with respect to translations and elastic distortions. However, the
disjoint nature of the pooling regions can limit generalization. Additionally,
as MP2-pooling reduces the size of the hidden layers so quickly, stacks
of back-to-back convolutional layers are needed to build really deep
networks \cite{NetworkInNetwork,VGG2014,GoogLeNet}. Two methods that
have been proposed to overcome this problems are:
\begin{itemize}
\item Using $3\times3$ pooling regions overlapping with stride 2 \cite{conf/nips/KrizhevskySH12}.
\item Stochastic pooling, where the act of picking the maximum value in
each pooling region is replaced by a form of size-biased sampling
\cite{StochasticPoolingZeilerFergus}.
\end{itemize}
However, both these techniques still reduce the size of the hidden
layers by a factor of two. It seems natural to ask if spatial-pooling
can usefully be applied in a gentler manner. If pooling was to only
reduce the size of the hidden layers by a factor of $\sqrt{2}$, then
we could use twice as many layers of pooling. Each layer of pooling
is an opportunity to view the input image at a different scale. Viewing
images at the `right' scale should make it easier to recognize
the tell-tale features that identify an object as belonging to a particular
class.

The focus of this paper is thus a particular form of max-pooling that
we call \emph{fractional max-pooling} (FMP). The idea of FMP is to
reduce the spatial size of the image by a factor of $\alpha$ with
$1<\alpha<2$. Like stochastic pooling, FMP introduces a degree of
randomness to the pooling process. However, unlike stochastic-pooling,
the randomness is related to the choice of pooling regions, not the
way pooling is performed inside each of the pooling regions.

In Section \ref{sec:Generalising-max-pooling} we give a formal description
of fractional max-pooling. Briefly, there are three choices that affect
the way FMP is implemented:
\begin{itemize}
\item The pooling fraction $\alpha$ which determines the ratio between
the spatial sizes of the input and the output of the pooling layer.
Regular $2\times2$ max-pooling corresponds to the special case $\alpha=2$.
\item The pooling regions can either be chosen in a \emph{random} or a \emph{pseudorandom}
fashion. There seems to be a trade off between the use of randomness
in FMP and the use of dropout and/or training data augmentation.
Random-FMP seems to work better on its own; however, when combined with `too much' dropout or training data augmentation,
underfitting can occur.
\item The pooling regions can be either \emph{disjoint} or \emph{overlapping}.
Disjoint regions are easier to picture, but we find that overlapping
regions work better.
\end{itemize}
In Section \ref{sec:Implementation} we describe how our convolutional
networks were designed and trained. In Section \ref{sec:Results}
we give results for the MNIST digits, the CIFAR-10 and CIFAR-100 datasets
of small pictures, handwritten Assamese characters and the CASIA-OLHWDB1.1
dataset of handwritten Chinese characters.

\begin{figure}
\begin{centering}
\includegraphics[height=2cm]{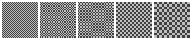}
\includegraphics[bb=78bp 0bp 115bp 40bp,clip,height=2cm]{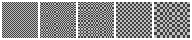}
\par\end{centering}

\caption{Left to right: A $36\times36$ square grid; disjoint pseudorandom
FMP regions with \mbox{$\alpha\in\{\sqrt[3]{2},\sqrt{2},2,\sqrt{5}\}$};
and disjoint random FMP regions for $\alpha=\sqrt{2}$. For $\alpha\in(1,2)$
the rectangles have sides of length 1 or 2. For $\alpha\in(2,3)$
the rectangles have sides of length 2 or 3. [Please zoom in if the images appear blurred.]\label{fig:pool}}
\end{figure}

\section{Fractional max-pooling\label{sec:Generalising-max-pooling}}

Each convolutional filter of a CNN produces a matrix of hidden variables.
The size of this matrix is often reduced using some form of pooling.
Max-pooling is a procedure that takes an $N_{\mathrm{in}}\times N_{\mathrm{in}}$
input matrix and returns a smaller output matrix, say $N_{\mathrm{out}}\times N_{\mathrm{out}}$.
This is achieved by dividing the $N_{\mathrm{in}}\times N_{\mathrm{in}}$
square into $N_{\mathrm{out}}^{2}$ pooling regions $(P_{i,j})$:
\[
P_{i,j}\subset\{1,2,\dots,N_{\mathrm{in}}\}^{2}\ \ \mathrm{for}\ \mathrm{each}\ \ (i,j)\in\{1,\dots,N_{\mathrm{out}}\}^{2},
\]
and then setting
\[
\mathrm{Output}_{i,j}=\max_{(k,l)\in P_{i,j}}\mathrm{Input}_{k,l}.
\]
For regular $2\times2$ max-pooling, $N_{\mathrm{in}}=2N_{\mathrm{out}}$
and $P_{i,j}=\{2i-1,2i\}\times\{2j-1,2j\}$. In \cite{conf/nips/KrizhevskySH12},
max-pooling is applied with overlapping $3\times3$ pooling regions
so $N_{\mathrm{in}}=2N_{\mathrm{out}}+1$ and the $P_{i,j}$ are $3\times3$
squares, tiled with stride 2. In both cases, $N_{\mathrm{in}}/N_{\mathrm{out}}\approx2$
so the spatial size of any interesting features in the input image
halve in size with each pooling layer. In contrast, if we take $N_{\mathrm{in}}/N_{\mathrm{out}}\approx\sqrt[n]{2}$
then the rate of decay of the spatial size of interesting features
is $n$ times slower. For clarity we will now focus on the case $N_{\mathrm{in}}/N_{\mathrm{out}}\in(1,2)$
as we are primarily interested in accuracy; if speed is an overbearing
concern then FMP could be applied with $N_{\mathrm{in}}/N_{\mathrm{out}}\in(2,3)$.

Given a particular pair of values $(N_{\mathrm{in}},N_{\mathrm{out}})$
we need a way to choose pooling regions $(P_{i,j})$. We will consider
two type of arrangements, overlapping squares and disjoint collections
of rectangles. In Figure \ref{fig:pool} we show a number of different
ways of dividing up a $36\times36$ square grid into disjoint rectangles.
Pictures two, three and six in Figure \ref{fig:pool} can also be
used to define an arrangement of overlapping $2\times2$ squares:
take the top left hand corner of each rectangle in the picture to
be the top left hand corner of one of the squares.

To give a formal description of how to generate pooling regions, let
$(a_{i})_{i=0}^{N_{\mathrm{out}}}$ and $(b_{i})_{i=0}^{N_{\mathrm{out}}}$
be two increasing sequence of integers starting at 1, ending with
$N_{\mathrm{in}}$, and with increments all equal to one or two (i.e.
$a_{i+1}-a_{i}\in\{1,2\}$). The regions can then be defined by either
\begin{equation}
P=[a_{i-1},a_{i}-1]\times[b_{j-1},b_{j}-1]\ \ \mathrm{or\ }\ P_{i,j}=[a_{i-1},a_{i}]\times[b_{j-1},b_{j}].\label{eq:pool}
\end{equation}
We call the two cases \emph{disjoint} and \emph{overlapping}, respectively.
We have tried two different approaches for generating the integer
sequence: using \emph{random} sequences of numbers and also using
\emph{pseudorandom }sequences.

\begin{figure}
\begin{centering}
\includegraphics[width=\columnwidth]{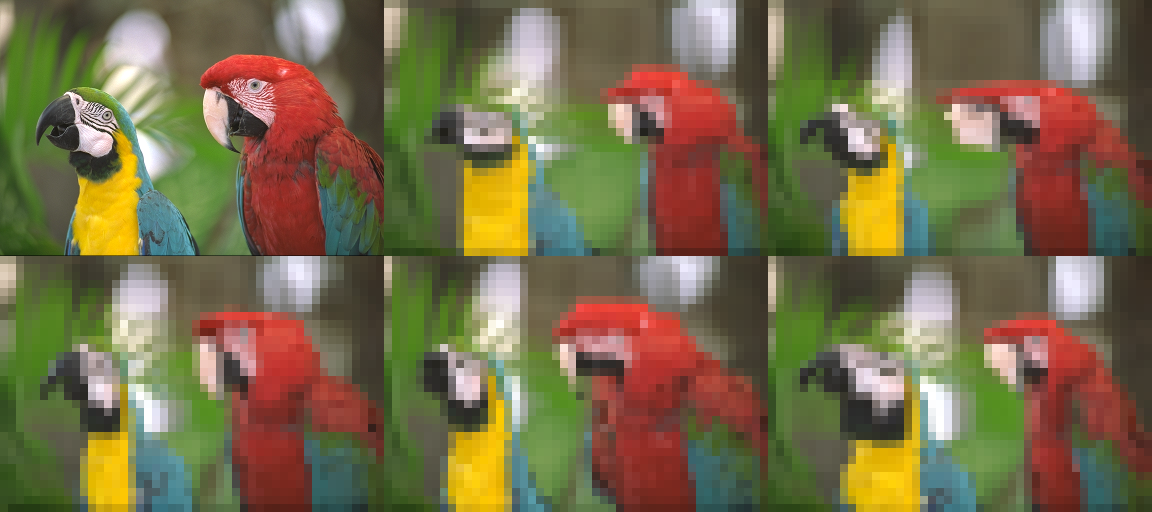}
\par\end{centering}

\caption{Top left, `Kodak True Color' parrots at a resolution of $384\times256$.
The other five images are one-eighth of the resolution as a result
of 6 layers of average pooling using disjoint random $\mathrm{FMP}\sqrt{2}$-pooling
regions. \label{fig:parrots}}
\end{figure}

We will say that the sequences are \emph{random} if the increments
are obtained by taking a random permutation of an appropriate number
of ones and twos. We will say that the sequences are \emph{pseudorandom}
if they take the form
\[
a_{i}=\mathrm{ceiling}(\alpha(i+u)),\qquad\alpha\in(1,2),\ \mathrm{with}\ \mathrm{some}\ u\in(0,1).
\]
 Below are some patterns of increments corresponding to the case $N_{\mathrm{in}}=25$,
$N_{\mathrm{out}}=18$. The increments on the left were generated\emph{
}`randomly', and the increments on the right come from pseudorandom
sequences:
\[
\begin{array}{cccc}
211112112211112122 &  &  & 112112121121211212\\
111222121121112121 &  &  & 212112121121121211\\
121122112111211212 &  &  & 211211212112121121
\end{array}
\]
Although both types of sequences are irregular, the pseudorandom sequences
generate much more stable pooling regions than the random ones. To
show the effect of randomizing the pooling regions, see Figure \ref{fig:parrots}.
We have taken a picture, and we have iteratively used disjoint random
pooling regions to reduce the size of the image (taking averages in
each pooling region). The result is that the scaled down images show
elastic distortion. In contrast, if we use pseudorandom pooling regions,
the resulting image is simply a faithfully scaled down version of
the original.

\section{Implementation\label{sec:Implementation}}

\begin{figure}
\begin{centering}
\includegraphics[scale=0.4]{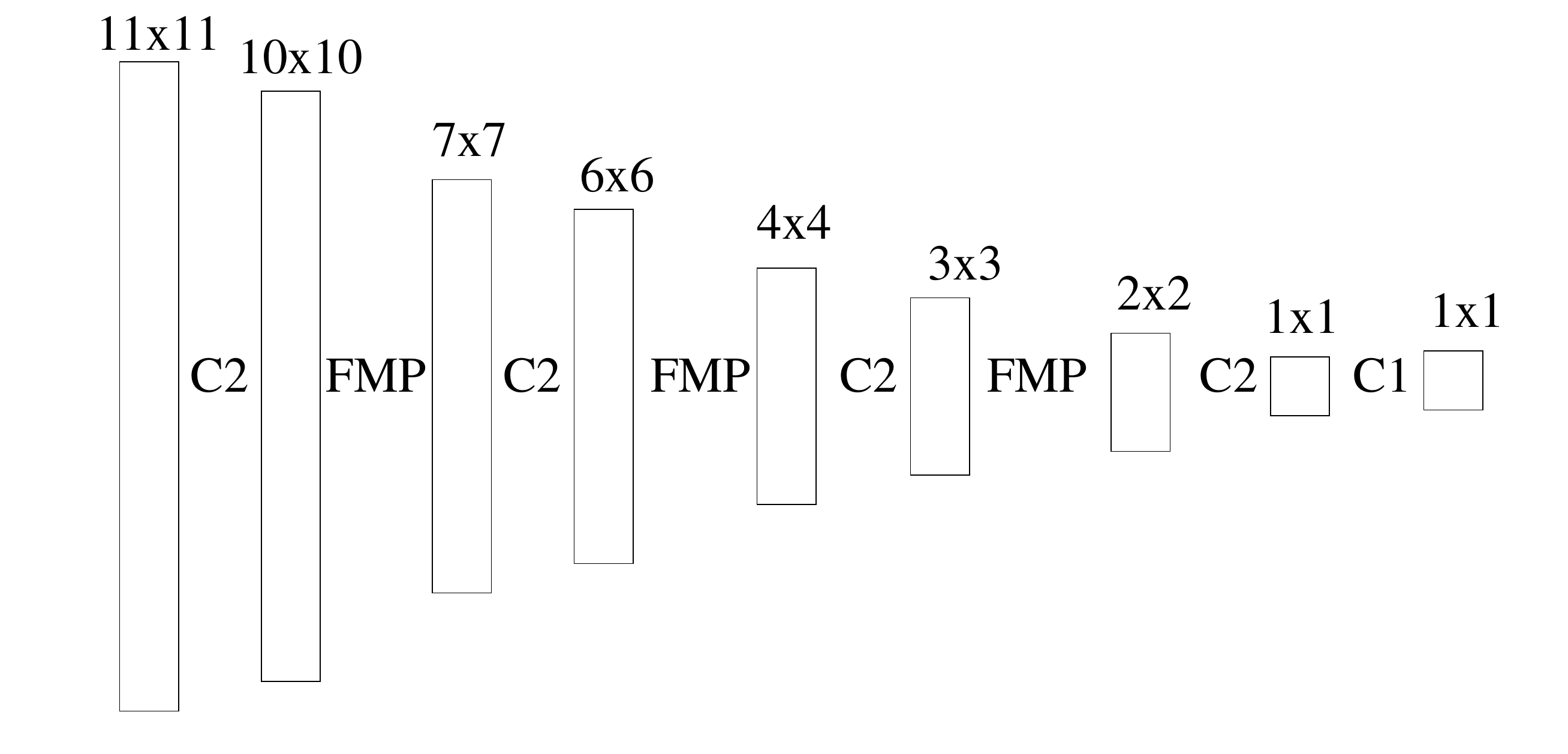}
\par\end{centering}

\caption{Layer sizes for a tiny FMP$\sqrt{2}$ network. The fractions $\tfrac{3}{2}$,
$\tfrac{6}{4}$ and $\tfrac{10}{7}$ approximate $\sqrt{2}$.\label{fig:Layer-sizes-for}}
\end{figure}

The networks are trainined using an implementation of a  sparse convolutional network \cite{GrahamSparse}.
What this means in practice is
that we can specify a convolutional network in terms of a sequence
of layers, e.g.
\[
10C2-FMP\sqrt{2}-20C2-FMP\sqrt{2}-30C2-FMP\sqrt{2}-40C2-50C1-\mathrm{output}.
\]
The spatial size of the input layer is obtained by working from right
to left: each C2 convolution increases the spatial size by one, and
FMP$\sqrt{2}$ layers increase the spatial size by a factor of $\sqrt{2}$,
rounded to the nearest integer; see Figure \ref{fig:Layer-sizes-for}.
The input layer will typically be larger than the input images---padding
with zeros is automatically added as needed. Fractional max-pooling
could also easily be implemented for regular convolutional neural
network software packages.

For simplicity, all the networks we use have a linearly increasing
number of filters per convolutional layer. We can therefore describe
the above network using the shorthand form
\[
(10nC2-FMP\sqrt{2})_{3}-C2-C1-\mathrm{output,}
\]
$10n$ indicates that the number of filters in the $n$-th convolutional
layer is $10n$, and the subscript 3 indicates three pairs of alternating
C2/FMP layers. When we use dropout, we use an increasing amount of
dropout the deeper we go into the network; we apply 0\% dropout in
the first hidden layer, and increase linearly to 50\% dropout in the
final hidden layer. We use leaky rectified linear units.

\subsection{Model averaging}

\begin{figure}
\begin{centering}
\includegraphics[scale=0.4]{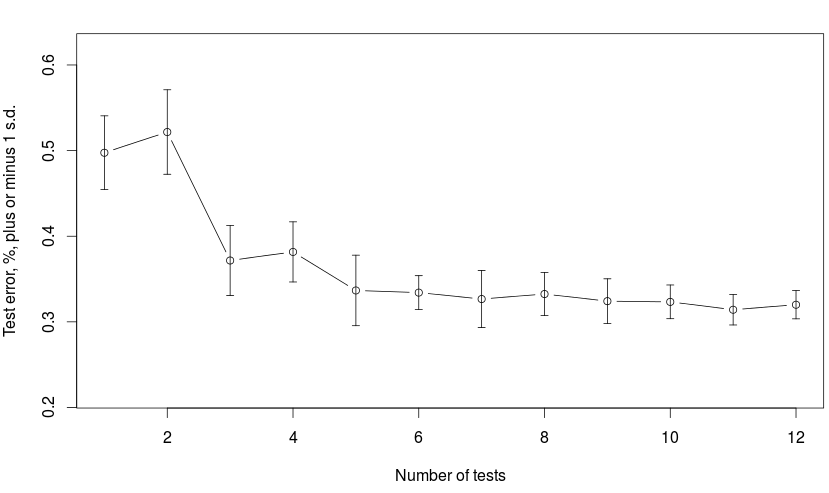}
\par\end{centering}

\caption{The effect of repeat testing for a single MNIST trained FMP network.
\label{jury-size}}
\end{figure}

Each time we apply an FMP network, either for training or testing
purposes, we use different random or pseudorandom sequences to generate
the pooling regions. An FMP network can therefore be thought of as
an ensemble of similar networks, with each different pooling-region
configuration defining a different member of the ensemble. This is
similar to dropout \cite{dropout}; the different values the dropout
mask can take define an ensemble of related networks. As with dropout,
model averaging for FMP networks can help improve performance. If
you classify the same test image a number of times, you may get a
number of different predictions. Using majority voting after classifying
each test image a number of times can substantially improve accuracy;
see Figure~\ref{jury-size}.

\section{Results\label{sec:Results}}

\subsection{Without training set augmentation or dropout\label{sec:withoutDataAug}}

To compare the different kinds of fractional max-pooling, we trained
FMP networks on the MNIST%
\footnote{\url{http://yann.lecun.com/exdb/mnist/}} set of digits and the CIFAR-100 dataset of small pictures \cite{CIFAR10}.
For MNIST we used a small FMP network:
\[
\mathrm{input}\ \mathrm{layer}\ \mathrm{size}\ 36\times36:\qquad(32nC2-FMP\sqrt{2})_{6}-C2-C1-\mathrm{output},
\]
and for CIFAR-100 we used a larger network:
\[
\mathrm{input}\ \mathrm{layer}\ \mathrm{size}\ 94\times94:\qquad(64nC2-FMP\sqrt[3]{2})_{12}-C2-C1-\mathrm{output}.
\]
Without using training data augmentation, state-of-the-art test errors
for these two datasets are 0.39\% and 34.57\%, respectively \cite{DeeplySupervisedNets}.
Results for the FMP networks are in Table \ref{tab:MNIST-and-CIFAR-10}.
Using model averaging with multiplicity twelve, we find that random
overlapping FMP does best for both datasets. For CIFAR-100, the improvement
over method using regular max-pooling is quite substantial.

To give an idea about network complexity, the CIFAR-100 networks have 12 million weights, and were trained for 250 repetitions of the training data (18 hours on a GeForce GTX 780). We experimented with changing the number of hidden units per layer for CIFAR-100 with random overlapping pooling:
\begin{itemize}
\item Using `$16nC2$' (0.8M weights) gave test errors of 42.07\% / 34.87\%.
\item Using `$32nC2$' (3.2M weights) gave test errors of 35.09\% / 29.66\%.
\item Using `$96nC2$' (27M weights) combined with dropout and a slower rate of learning rate decay gave test errors of 27.62\% / 23.82\%.
\end{itemize}

\begin{table}
\centering{}
\begin{tabular}{|c|cccc|}
\hline
Dataset and the number & pseudorandom & random & pseudorandom  & random\\
of repeat tests & disjoint & disjoint & overlapping & overlapping\\
\hline
MNIST, 1 test & 0.54 & 0.57 & \textbf{0.44} & 0.50\\
MNIST, 12 tests & 0.38 & 0.37 & 0.34 & \textbf{0.32}\\
CIFAR-100, 1 test & 31.67 & 32.06 & \textbf{31.2} & 31.45\\
CIFAR-100, 12 tests & 28.48 & 27.89 & 28.16 & \textbf{26.39}\\
\hline
\end{tabular}
\ \\
\ \\
\caption{MNIST and CIFAR-100 \% test errors.\label{tab:MNIST-and-CIFAR-10} }
\end{table}

\subsection{Assamese handwriting }

To compare the effect of training data augmentation when using FMP
pooling versus MP2 pooling, we used the The Online Handwritten Assamese
Characters Dataset \cite{UCIrep}. It contains 45 samples for each
of 183 Indo-Aryan characters. `Online' means that each pen stroke is represented as a sequence of $(x,y)$ coordinates.
We used the first 36 handwriting samples
as the training set, and the remaining 9 samples for a test set. The
characters were scaled to fit in a box of size $64\times64$. We trained
a network with six layers of $2\times2$ max pooling,
\[
32nC3-MP2-(C2-MP2)_{5}-C2-\mathrm{output}
\]
and an FMP network using 10 layers of random overlapping FMP$\sqrt{2}$
pooling,
\[
(32nC2-FMP\sqrt{2})_{10}-C2-C1-\mathrm{output}.
\]
We trained the networks without dropout, and either
\begin{itemize}
\item no training data augmentation,
\item with the characters shifted by adding random translations, or
\item with affine transformations, using a randomized mix of translations,
rotations, stretching, and shearing operations.
\end{itemize}
See Table \ref{tab:Assamese-=000025-test}. In a sense, max-pooling
and training data augmentation are two different ways of encoding
our apriori knowledge that the meaning of handwriting is generally
invariant under certain kinds of minor distortions. Interestingly,
the FMP network without data augmentation does better than the MP2
network with training data augmentation, suggesting that FMP is a
better way of encoding that information.

\begin{table}
\centering{}%
\begin{tabular}{|c|ccc|}
\hline
Pooling method & None & Translations & Affine\\
\hline
6 layers of MP2 & 14.1 & 4.6 & 1.8\\
10 layers of FMP($\sqrt{2})$, 1 test & 1.9 & 1.3 & 0.9\\
10 layers of FMP($\sqrt{2})$, 12 tests & \textbf{0.7} & \textbf{0.8} & \textbf{0.4}\\
\hline
\end{tabular}
\ \\
\ \\
\caption{Assamese \% test error with different type of data augmentation.\label{tab:Assamese-=000025-test}}
\end{table}

\subsection{Online Chinese handwriting}

The CASIA-OLHWDB1.1 database contains online handwriting samples of
the 3755 isolated GBK level-1 Chinese characters \cite{CASIA}. There are approximately
240 training characters, and 60 test characters, per class. A test
error of 5.61\% is achieved using 4 levels of MP2 pooling \cite{multicolumndeep}.

We used the representation for online characters described in \cite{GrahamSparse}; the characters were drawn with size $64\times64$ and additional features measuring the direction of the pen are added to produce an array of size $64 \times 64 \times 9$. Using 6 layers of $2\times2$ max-pooling, dropout and affine training data augmentation
resulted in a 3.82\% test error \cite{GrahamSparse}.
Replacing max-pooling
with pseudorandom overlapping FMP:
\[
(64nC2-FMP\sqrt{2})_{7}-(C2-MP2-C1)_{2}-C2-C1-\mathrm{output}
\]
results in test errors of 3.26\% (1 test) and 2.97\% (12 tests).

\subsection{CIFAR-10 with dropout and training data augmentation}
For CIFAR-10 we used dropout and extended the training set using affine transformations: a randomized mix of translations,
rotations, reflections, stretching, and shearing operations.
We also added random shifts to the pictures in RGB colorspace. For a final 10 training epochs, we trained the network without the affine transformations.

For comparison, human performance on
CIFAR-10 is estimated to be 6\%\footnote{\url{http://karpathy.ca/myblog/?p=160}}.
A recent Kaggle competition relating to CIFAR-10 was won with a test
error of 4.47\%\footnote{\url{https://www.kaggle.com/c/cifar-10/}} using the same training data augmentation scheme, and architecture
\[
(300nC2-300nC2-MP2)_5-C2-C1-\mathrm{output}.
\]
Using a pseudorandom overlapping pooling FMP network
\[
(160nC2-FMP\sqrt[3]{2})_{12}-C2-C1-\mathrm{output}.
\]
we obtained test errors of
4.50\% (1 test), 3.67\% (12 tests) and 3.47\% (100 tests).


\section{Conclusions}

We have trained convolutional networks with fractional max-pooling
on a number of popular datasets and found substantial improvements
in performance. Overlapping FMP seems to be better than disjoint FMP.
Pseudorandom pooling regions seem to do better than random pooling
regions when training data augmentation is used. It is possible that random pooling might
regain the upperhand if we fine-tuned the amount of dropout used.

Looking again at the distortions created by random pooling in Figure
\ref{fig:parrots}, note that the distortion is `decomposable' into
an $x$-axis distortion and a $y$-axis distortion. It might be interesting
to explore pooling regions that cannot be written using equation \ref{eq:pool},
as they might encode more general kinds of distortion into the resulting
convolutional networks.

\end{document}